\documentclass{article}
\usepackage{arxiv}

\usepackage{hyperref}
\usepackage{bm}% allocate bold fonts
\usepackage{cite}
\usepackage{graphics,fancyhdr,graphicx,subfigure}
\usepackage{subfigure}
\usepackage{amsthm,amsmath,amsfonts,latexsym,amssymb}
\usepackage{lineno}
\usepackage{diagbox}
\usepackage{algorithm}
\PassOptionsToPackage{noend}{algpseudocode}
\usepackage{algpseudocode}
\usepackage{multirow,booktabs}
\usepackage{listings}
\usepackage[table]{xcolor}
\usepackage{lscape}
\usepackage{comment}
\usepackage{tabularx}
\usepackage{graphicx}%to add images
\usepackage{tabularx}
\usepackage{arydshln}
\usepackage{nomencl}%for adding symbols page
\usepackage{array}%to add table
\usepackage{caption}%add caption for the table and list it in list of tables
\usepackage{breqn}
\usepackage{lineno}
\usepackage{mathtools}
\usepackage{subfigure}
\usepackage{caption}
\usepackage{lmodern}
\usepackage{calligra}
\usepackage{xspace}
\usepackage[utf8]{inputenc}%skiplength
\usepackage{enumerate}%list
\usepackage[english]{babel}
\def\equationautorefname~#1\null{%
  Eq.~(#1)\null
  }
\def\subfigureautorefname~#1\null{%
  Fig.~#1\null
}

\definecolor{listinggray}{gray}{0.9}
\definecolor{lbcolor}{rgb}{0.9,0.9,0.9}
\definecolor{Darkgreen}{RGB}{0,100,0}

\title{Waveformer for modelling dynamical systems \thanks{https://www.csccm.in/}}

\author{ \hspace{1mm}Navaneeth~N.\\
	Department of Applied Mechanics\\
	Indian Institute of Technology (IIT) Delhi\\
	Hauz Khas - 110 016, New Delhi, India \\
	\texttt{navaneeth.n@am.iitd.ac.in} \\
	\And
	\hspace{1mm}Souvik~Chakraborty \\
	Department of Applied Mechanics\\
        Yardi School of Artificial Intelligence \\
	Indian Institute of Technology (IIT) Delhi\\
	Hauz Khas - 110 016, New Delhi, India \\
	\texttt{souvik@am.iitd.ac.in} \\
}

\renewcommand{\shorttitle}{Waveformer for modelling dynamical systems}

\hypersetup{
pdftitle={Waveformer for modelling dynamical systems},
pdfsubject={q-bio.NC, q-bio.QM},
pdfauthor={Navaneeth N., Souvik Chakraborty},
pdfkeywords={Operator learning, Wavelet transformations, Transformers, Dynamical systems},
}
\begin{document}
\maketitle

\begin{abstract}
Neural operators have gained recognition as potent tools for learning solutions of a family of partial differential equations. The state-of-the-art neural operators excel at approximating the functional relationship between input functions and the solution space, potentially reducing computational costs and enabling real-time applications. However, they often fall short when tackling time-dependent problems, particularly in delivering accurate long-term predictions. In this work, we propose ``waveformer'', a novel operator learning approach for learning solutions of dynamical systems. The proposed waveformer exploits wavelet transform to capture the spatial multi-scale behavior of the solution field and transformers for capturing the long horizon dynamics. We present four numerical examples involving Burgers’s equation, KS-equation, Allen Cahn equation, and Navier Stokes equation to illustrate the efficacy of the proposed approach. Results obtained indicate the capability of the proposed waveformer in learning the solution operator and show that the proposed Waveformer can learn the solution operator with high accuracy, outperforming existing state-of-the-art operator learning algorithms by up to an order, with its advantage particularly visible in the extrapolation region 
\end{abstract}

\keywords{
Operator learning\and Wavelet transformations\and Transformers\and Dynamical systems}

\section{Introduction}\label{sec:intro}
Modeling physical and real-world phenomenons, ranging from fluid dynamics to electromagnetism, as a system of Partial Differential Equations (PDEs) is a prevalent approach in the scientific domain. While analytical solutions are possible for a limited class of problems, the majority of real-world problems necessitate the use of numerical methods. In general, numerical methods \cite{kang1996finite,brenner2004finite,cottrell2009isogeometric, ozicsik2017finite, eymard2000finite} involve discretizing the domain of the problem and approximating the solution at discrete points. These numerical methods often result in exorbitant computational costs \cite{thompson2006review} and prohibit the implementation in real-time applications. Consequently, the development of efficient methods for solving PDEs remains a relevant area of computational science and engineering research.

Surrogate-based approaches have recently gained attention due to the emergence of data-driven methods and scientific machine learning methods (Sci-ML methods). These ML-based approaches have proven to be effective tools for mitigating the computational bottlenecks associated with state-of-the-art numerical methods. There is a wealth of literature available that delves into the insights and the details of the implementation of the Sci-ML-based surrogate modeling methods such as polynomial chaos expansions \cite{blatman2011adaptive,crestaux2009polynomial}, Gaussian processes \cite{schulz2018tutorial,deringer2021gaussian}, reservoir computing \cite{lukovsevivcius2009reservoir,chattopadhyay2020data} and deep neural networks \cite{montufar2014number, szegedy2013deep,arisoy2012deep,bar2019learning}. Although the purely data-driven models efficiently approximate the physical phenomena in the trained regime, the predictions often diverge significantly beyond the training window. Moreover, the models do not guarantee the underlying physics is satisfied. Ensuing research to overcome this shortcoming has resulted in a new paradigm of learning PDE solutions, known as physics-informed learning \cite{raissi2017physics, yang2018physics,navaneeth2023stochastic}, where the models are trained by incorporating the physics into the loss function. Certainly, the learning methodologies of data-driven neural networks and physics-informed neural networks are different. However, physics-informed algorithms are notoriously hard to train. More importantly, the network needs to be trained every time a system parameter or the input condition changes.

Operator learning refers to a class of machine learning techniques that involves using neural networks to learn operators or functional mapping between inputs and outputs. The approach has proven to be promising in learning the solution of the family of parametric PDEs, where a single operator enables the mapping between the input function space and the solution space. Thus, in essence, the training of the framework once learns the solutions of PDEs for the varying input conditions and system parameters that were hitherto unattainable with conventional data-driven and physics-informed neural networks. DeepoNet \cite{lu2021learning, wang2023long} is a neural operator framework that was among the first of its kind to be built on the universal approximation theorem for operators. Furthermore, a novel approach is introduced, reinterpreting operator learning as the process of learning the parameterized kernels of integral transformations from the paired input functions and corresponding solution space. Examples of such operator learning paradigms include Graph Neural Operators (GNO) \cite{anandkumar2020neural,li2020multipole}, Fourier Neural Operators (FNO) \cite{li2020fourier, kovachki2021universal}, and Wavelet Neural Operators (WNO) \cite{tripura2022wavelet, navaneeth2023physics,gupta2021multiwavelet}. 
While in GNO, the kernel of integral transforms is learned through a message-passing interface between the graph networks, FNO learns the parameters of the integral kernel in Fourier space, where spectral decomposition enables the framework to obtain the integral kernel. FNO outperforms most state-of-the-art methods. However, it is liable to a significant limitation on its spatial resolution capabilities, attributed to the frequency-localized nature of the basis functions inherent in the Fast Fourier Transform (FFT). This often leads to suboptimal performance of FNO for learning the solutions of the PDEs with complex geometry or capturing spatial characteristics of signals. On the contrary, WNO remains impervious to this limitation as it utilizes wavelet transforms and exploits them to learn the variation in the input patterns over spatial coordinates through spatial and frequency localization. The wavelet transform decomposes the input signal into wavelet coefficients at multiple scales \cite{zhang2019wavelet}, with each scale corresponding to a different frequency band. This allows WNOs to capture the behavior of the solution in problems that involve multi-scale dynamics.

As was mentioned previously, the operator learning paradigms excel at learning solution operators; however, their effectiveness in tasks such as learning solutions for dynamical systems is limited, primarily due to the absence of mechanisms for preserving long-term temporal dynamics. To that end, we propose a new operator learning paradigm called ``waveformer''; it combines elements of wavelet transformation, kernel integral operator, and transformer technology. Transformers have showcased remarkable performance in natural language processing \cite{wolf2020transformers,arkhipov2019tuning} and sequence-to-sequence learning tasks \cite{hawthorne2021sequence,dong2018speech,lim2021temporal}. Their key strength lies in their ability to capture dependencies between elements within a sequence through a specialized feature known as the attention mechanism \cite{wiegreffe2019attention}. The proposed waveformer harnesses the transformer's strengths in capturing temporal evolution and combines it with the power of wavelet decomposition to effectively learn the spatial variations of the response.

The remainder of the paper is organized as follows. The general problem statement is described in Section \ref{sec:Problem statement}. The proposed approach is elucidated in the following Section \ref{sec: Methedology}. Subsequently, Numerical examples are presented in the section\ref{sec: Numerical example}. Finally, the conclusion and final notes are provided in Section \ref{sec:conclusion}.

\section{Background theory}\label{sec:Problem statement}
\subsection{Operator learning}
The methodology essentially aims to learn the functional mapping between the infinite-dimensional input and the output function spaces. To delve into mathematical formulations, we consider the function spaces, $\mathcal{A}$ and $\mathcal{U}$, contain a collection of all the inputs $\bm{a} \in \mathcal{A}$ and outputs $\bm{u} \in \mathcal{U}$. Now, we assume that a nonlinear operator $\mathcal{D}$ enables the mapping between separable function spaces $\mathcal{A}$ and $\mathcal{U}$, $\mathcal{D}: \mathcal{A} \mapsto \mathcal{U}$. In order to approximate the operator, a machine learning model can be conveniently employed, provided that we have access to the $N$ number of paired observations of inputs and outputs $\{\bm{a}_{j},\bm{u}_{j}\}^{N}_{j=1}$.
\begin{equation}\label{neural operator}
    \mathcal{D}:\mathcal{A}\times \bm{\theta}_{NN} \mapsto \mathcal{U},  
\end{equation}
where $\bm{\theta}_{NN}$ denotes the trainable parameter space of the operator devised for the neural network. The operator learning assumes a unique solution $\bm{u} = \bm{\theta}_{NN}(\bm{a}) \in \mathcal{U}$, exists for any $\bm{a} \in \mathcal{A}$. Thus, with $N$ samples of paired input-output data, the data-driven neural operator can be trained by minimizing the loss   $\mathcal{L}_{\text {data }}$, which is expressed as:
\begin{equation}
    \mathcal{L}_{\text {data }}\left(\bm{u}, \bm{\theta}_{NN}(\bm{a})\right)=\left\|\bm{u}-\bm{\theta}_{NN}(\bm{a})\right\|_{\mathcal{U}}^2=\int_D\left|u(x_i)-\bm{\theta}_{NN}(a)(x_i)\right|^2\mathrm{~d} x
\end{equation}
Averaging error across all possible inputs, and with domain discretizations of $n_d$, the operator loss can be computed as:
\begin{equation}
    \mathcal{L}_{\text {data }}\left(\bm{u}, \bm{\theta}_{NN}(\bm{a})\right) = \frac{1}{N} \sum_{j=1}^N \sum_{i=1}^{n_d}\left|u_j(x_i)-\mathcal{D}\left(a_j, \bm{\theta}_{NN} \right)(x_i)\right|^2 
\end{equation}
Finally, minimizing the loss function yields optimal network parameters:
\begin{equation}
    \bm{\theta}^{*}_{NN} = \underset{\bm{\theta}_{NN}} {\text{argmin}}\; \mathcal{L}_{\text {data }}\left(\bm{u}, \bm{\theta}_{NN}(\bm{a})\right)
\end{equation}

We now focus on a specific class of problem, where the goal is to learn the solution of a family of parametric partial differential equations (PDEs). Within  the function spaces, a family of parametric PDEs $\mathcal{N}$ takes the form:
\begin{equation}\label{operator1}
    \mathcal{N}(\bm{a},\bm{u}) = \bm{0}, \,\, \text { in } D \subset \mathbb{R}^d.
\end{equation} 
Here is the differential operator corresponding to the parametric PDE,  where the PDE is defined on an $d$-dimensional bounded domain, $D \in \mathbb{R}^{d}$. The domain is defined by a boundary $\partial D$, where the boundary condition is expressed as:
\begin{equation}\label{operator_BC}
    \bm{u} = \bm{g}, \,\, \text { in } \partial D .
\end{equation}
The parameter $\bm{a} \in \mathbb{R}^{a}$ denotes the input function space, and $\bm{u} \in \mathbb{R}^{u}$ denotes the solution space of the parametric PDE. For a fixed domain $x \in D$, the input function space of the operator contains the source term 
$f(x, t): D \mapsto \mathbb{R}$, the initial condition $u(x, 0): D \mapsto \mathbb{R}$, and the boundary conditions $u(\partial D, t): D \mapsto \mathbb{R}$, while the output function space comprises the solution of the parametric PDE, $u(x, t): D \mapsto \mathbb{R}$, with $t$ being the time coordinate. Our interest is in learning the integral operator $\mathcal{D}: \mathcal{A} \mapsto \mathcal{U}$, corresponding to the differential operator $\mathcal{N}$, which maps the input functions to the solution space. 

\subsection{Kernel-based neural Operator theory}
A generalized Hammerstein integral equation for a nonlinear parametric PDE is given as:
\begin{equation}\label{eq:hammerstein}
    u(x) = \int_{D} k(x, \xi)f\left(\xi, u(\xi)\right) d\xi + g(x); \quad x \in D,
\end{equation}
where $k(\cdot)$ denotes the kernel of the nonlinear integral equation and which can be interpreted as the nonlinear counterpart of Green's function, and $g$ represents some nonlinear transformation. We note here that the function $f(x,s)$ is a given continuous function that satisfies the condition (Lipschitz criteria):
\begin{equation}
    |f(x,s)|\leq C_1|s|+C_2
\end{equation}
where $C_1$ and $C_2$ are positive constants such that  $C_1$ is smaller than the first eigen value of the kernel $K(x,s)$. Now to learn the nonlinear integral operator, $\mathcal{D}: a(x) \mapsto u(x)$, we use the expression of integral provided in \autoref{eq:hammerstein}. Further, we discretize the solution domain $D \in \mathbb{R}^{d}$ to obtain a finite-dimensional parameterized space, as the integral formulation in \autoref{eq:hammerstein} is defined only on a finite-dimensional space. A local transformation ${{P}}$ is employed to incorporate the multi-dimensional kernel convolution, where  ${\rm{P}}$ maps the input $a(x)$ to a higher-dimensional space ${d_v}$. In a neural network-based architecture, the local transformation ${{P}}$ can be implemented using either a fully connected neural network layer (FNN) or a convolution with a kernel size $1\times1$. Subsequently, $m$-number of iterations of  \autoref{eq:hammerstein} are performed on the lifted space. An 
operation $G:\mathbb{R}^{d_v} \mapsto \mathbb{R}^{d_v}$  enables the transformation such that $v_{i+1} = G(v_{i})$. Followed by the $m$-iterations, a second local transformation ${{Q}}: v_{m}(x) \mapsto u(x)$ is performed to achieve the final solution space $u(x) \in \mathbb{R}^{d_u}$. The step-wise updation though the composition operator  $G(\cdot)$ can be defines as:
\begin{equation}\label{eq:iteration}
    v_{i+1} = G(v_{i})(x):= \varphi \left( \left(K(a; \phi) * v_{i}\right)(x) + W v_{i}(x) \right); \quad x \in D, \quad i \in [1,m].
\end{equation}
Here $\varphi(\cdot) \in \mathbb{R}$ represents a non-linear activation function. While $K$ is the nonlinear integral operator parameterized by  the trainable neural network, $\phi \in \theta_{NN}$, $W: \mathbb{R}^{d_{v}} \to \mathbb{R}^{d_{v}}$ denotes a linear transformation. In essence, the integral operator $K$ can be reinterpreted in terms of  the kernel of the nonlinear integral equation in \autoref{eq:hammerstein} as:
\begin{equation}\label{eq:integral}
    \left(K(a ; \phi) * v_{j}\right)(x) := \int_{D} k \left(a(x), x, \xi; \phi \right) v_{j}(\xi) \mathrm{d}\xi; \quad x \in D, \quad j \in [1,l].
\end{equation}
As we discussed in the \autoref{sec:intro}, FNO and WNO are operators devised to lean kernel parameters in the transformed domain. While in the FNO, the convolution operator defined in Fourier space where higher-order Fourier modes are truncated, WNO learns the kernel $k \left(a(x), x, \xi; \phi \right)$ by parameterizing the neural network in the wavelet space.

\subsection{Wavelet transformation}
Wavelets offer numerous advantages in comparison with the traditional Fourier methods \cite{daubechies1992ten}. Primarily, wavelets can effectively capture and analyze localized features in signals as they are localized in both spatial and frequency. Secondly, wavelets possess a structured hierarchical nature; thus, in reconstruction using the inverse wavelet transform, the learned parameters propagate automatically to lower decomposition levels. This ensures the preservation of knowledge and facilitates effective learning. It is noteworthy that there exist, diverse families of wavelets, each exhibiting unique characteristics. While the proposed framework can be applied to any wavelet family, we specifically focus on the Daubechies wavelets. The Daubechies wavelets are known for their simplicity, compactness, orthogonality, and resilience in texture retrieval tasks. Additionally, they offer the advantage of adaptability to different levels of smoothness as per the specific requirements \cite{daubechies1992ten}.

The forward and inverse wavelet transforms, denoted as $\mathcal{W}(\cdot)$ and $\mathcal{W}^{-1}(\cdot)$ respectively, are defined according to the formulation by Daubechies \cite{daubechies1992ten}. These transforms can be represented as follows:
\begin{equation}\label{eq:wavelet}
    \begin{aligned}
        (\mathcal{W} v_{j})(s, \tau) & = \int_{D} \Gamma (x) \frac{1}{|s|^{1 / 2}} \psi\left(\frac{x-\tau}{s}\right) dx, \\
        (\mathcal{W}^{-1} (v_{j})_w)(x) & = \frac{1}{C_{\psi}} \int_{0}^{\infty} \int_{D} (v_{j})_{w}(s, \tau) \frac{1}{|s|^{1 / 2}} \tilde{\psi}\left(\frac{x-\tau}{s}\right) d\tau \frac{ds}{s^{2}},
    \end{aligned}
\end{equation}
In the above equations, $\psi(x)$ represents the orthonormal mother wavelet, while $s$ and $\tau$ are the scaling and translational parameters involved in wavelet decomposition. The term $(v_{j})_{w}$ denotes the wavelet decomposed coefficients of $v{j}(x)$, and $\psi(\cdot)$ represents the scaled and shifted mother wavelet and the admissible constant $0 < C_{\psi} < \infty$ is introduced as defined by Daubechies \cite{daubechies1992ten}. Here we note that in the wavelet neural operator, the kernel integration $R_{\phi} = \mathcal{W}(k_{\phi})$ is performed through convolution in the wavelet domain and is denoted by:
\begin{equation}\label{eq:conv_final}
\begin{aligned}
\left(K(\phi) * v_{i}\right)(x)=\mathcal{W}^{-1}\left(R_{\phi} \cdot \mathcal{W}( v_{i})\right)(x); && x \in D.
\end{aligned}
\end{equation}

\section{Proposed framework:Waveformer}\label{sec: Methedology}
The core of kernel-based neural operator learning is the step-wise updation given by the equation \autoref{eq:iteration}, where kernel of the integral transformation is parameterized by a neural network. 
Waveformer introduces a novel architecture that redefines the conventional operator learning architecture, where the transformer replaces the stage involving the action of the nonlinear function and step-wise updation. More precisely, the waveformer performs nonlinear integration on wavelet-transformed uplifted input using a transformer. The operation is given by:
\begin{equation}\label{eq:trans1}
    v_{out1}(x) = \mathcal{W}^{-1}(T_{\mathcal{W}}(\mathcal{W}(v_{in}(x))))
\end{equation}
where $v_{in}$ represents the uplifted input, $T_{\mathcal{W}}$ denotes the transformer acts on the wavelet domain and $v_{out1}$ is the corresponding output. Moreover, the action of transformer $T_{R}$ yields a nonlinear transformation on the physical space, which in effect replaces the series of activations and linear transformations in the \autoref{eq:iteration}, which is given by;
\begin{equation} \label{eq:trans2}
    v_{out2}(x) = T_{\mathcal{R}}(v_{in}(x))
\end{equation}    
Here $v_{out2}(x)$ is the output of the transformer $T_{R}$. Hence, the output of the overall composition operation of the waveformer can be expressed as the summation of the outputs $v_{out1}$ and $v_{out2}$.
\begin{equation}\label{eq:summ}
    v_{out} = G_{T}(v_{in}) = {\mathcal{W}^{-1}}{(T_{\mathcal{W}})}{({v_{in}}(x))}+T_{R}({v_{in}}(x))
\end{equation}
where $\mathcal{W}(\cdot)$ and $\mathcal{W}^{-1}(\cdot)$ denote the forward and inverse wavelet transforms, respectively, are defined according to the formulation by Daubechies \cite{daubechies1992ten}.
Here we note that diverse wavelet families exist, each exhibiting unique characteristics. While the proposed framework can be applied to any wavelet family, we specifically focus on the Daubechies wavelets in the current work. An algorithm describing the implementation of the waveformer is provided in algorithm \ref{algo:Waveformer}. 
\begin{algorithm}[ht!]
    \caption{Algorithm for waveformer}\label{algo:Waveformer}
    \begin{algorithmic}[1]
    \Require{$N$-number of paired input and output $\left\{a(x) \in \mathbb{R}^{n_D \times d_a}, u(x) \in \mathbb{R}^{n_D \times d_u}\right\}$ and coordinates $x \in D$}
         \State {\textbf{Initialize:} Network parameters, $\bm{w} = \{w_{i}s, b_{i}s\}$} of the waveformer, $\bm{\theta}_{NN}$.
        \State{Stack the inputs: $\{a(x), x\} \in \mathbb{R}^{n_D \times 2 d_a}$.}
        \For{for epoch $=1, \ldots$, epochs}
            \State{Perform the uplift transformation $P(\cdot): v_{in}(x) \in \mathbb{R}^{n_D \times d_v}=P\left(\{a(x), x\} \in \mathbb{R}^{n_D \times 2 d_a}\right)$}
            
            \State{Wavelet decomposition of the uplifted input up to $p$ levels: $\mathcal{W}\left(v_{in}(x)\right) \in \mathbb{R}^{n_D / 2^p \times d_v}$.}
            
            \State {Employment of the transformer $T_{\mathcal{W}}$ in the wavelet space: $v_{out1}(x) = T_{\mathcal{W}}(\mathcal{W}(v_{in}(x)))$}
            
            \State {Transform back to physical domain:$ v_{out1}(x) = \mathcal{W}^{-1}(T_{\mathcal{W}}(\mathcal{W}(v_{in}(x))))$}\label{step:7}  \Comment{Eq. \eqref{eq:trans1}}
            
            \State{Action of transformer $T_{R}$ on $v_{in}$  : $v_{out2}(x) = T_{\mathcal{R}}(v_{in}(x))$} \label{step:8}  \Comment{Eq. \eqref{eq:trans2}}
            
            \State{The outputs of the step \ref{step:7} and \ref{step:8} are summed to obtain uplifted output $v_{out} \in \mathbb{R}^{n_D \times d_v}$ } \Comment{Eq. \eqref{eq:summ}}
            
            \State{Final out $\hat{u(x)} \in \mathbb{R}^{n_D \times  d_u}$ through an activated down lifting using transformation $Q(\cdot) :\hat{u(x)} = Q(v_{out})$} 
            
            \State{Compute the reconstruction loss:$\mathcal{L}(${u(x)},$\hat{u(x)})$}
            
            \State{Updating of the network parameters : 
             $\{w_{i}s, b_{i}s\}\leftarrow \{w_{i}s, b_{i}s\}-\delta \nabla_{\bm w, \bm b}\mathcal{L}(\bm w, \bm b)$}
        \EndFor
    \Ensure{Output prediction of waveformer $u \in \mathcal{U}$ for the optimal network parameters}
    \end{algorithmic}
\end{algorithm}

\subsection{Attention and kernel integration}
Self-attention and encoder-decoder cross-attention are the vital mechanisms that allow the transformer to learn temporal dependencies. The most commonly used attention model is the scaled-dot product where the key, query, and value embedding generated for an input tensor $Y \in \mathbb{R}^{d}$ is given by:
\begin{equation}
    \bm{q} = Y W_q, \quad \bm{k} = Y W_k, \quad \bm{k}, \quad \bm{v} = Y W_v
\end{equation}
where $W_q, W_k, W_v \in \mathbb{R}^{d \times d}$ are the query, key, and value matrices, respectively. When the input has sequence length on N, the corresponding attention embedding  through the self-attention, $\mathbb{R}^{N \times d} \rightarrow \mathbb{R}^{N \times d}$ is given by:
\begin{equation}
    \mathcal{A}(Y):=\operatorname{softmax}\left(\frac{Y W_q\left(Y W_k\right)^{\top}}{\sqrt{d}}\right) Y W_v
\end{equation}
As suggested by the works \cite{tsai2019transformer,cao2021choose,guibas2021adaptive} the self-attention mechanism can be reinterpreted as a kernel integration. While the scalar-valued $K$ is defined as $K:=\operatorname{softmax}\left(\left\langle Y W_q, Y W_k\right\rangle / \sqrt{d}\right)$, the self-attention embedding in terms of the kernel, $\kappa[s, t]$ parameterized by $[s,t]$ can be obtained as: 
\begin{equation}
    \mathcal{A}(Y)[s]:=\sum_{t=1}^N Y[t] \kappa[s, t] \quad \forall s \in[N] .
\end{equation}
where the parameterised kernel $\kappa[s, t]=K[s, t]$ expressed in terms of K as:
\begin{equation}
    \kappa[s, t]=K[s, t]\cdot W_v
\end{equation}
Further kernel summation can be extended to continuous kernel integrals.

The transformer architecture is comprised of a series of stacked encoder and decoder blocks. While the encoder utilizes self-attention, the decoder utilizes both self-attention and encoder-decoder cross-attention. In encoder-decoder cross-attention, the attention scores corresponding to attention embedding are computed between the query of the decoder input and the keys of the encoder embedding. The encoder-decoder cross-attention mechanism enables the decoder to leverage the encoded information from the input sequence, aligning it with the generation process and allowing the model to produce accurate and contextually relevant outputs. In summary, the self-attention mechanism in the transformer facilitates continuous kernel integration within the wavelet domain, and the encoder-decoder attention enables the framework to preserve temporal information.

\subsection{Training waveformer}
\begin{figure}[t!]
    \centering
    \includegraphics[width=\textwidth]{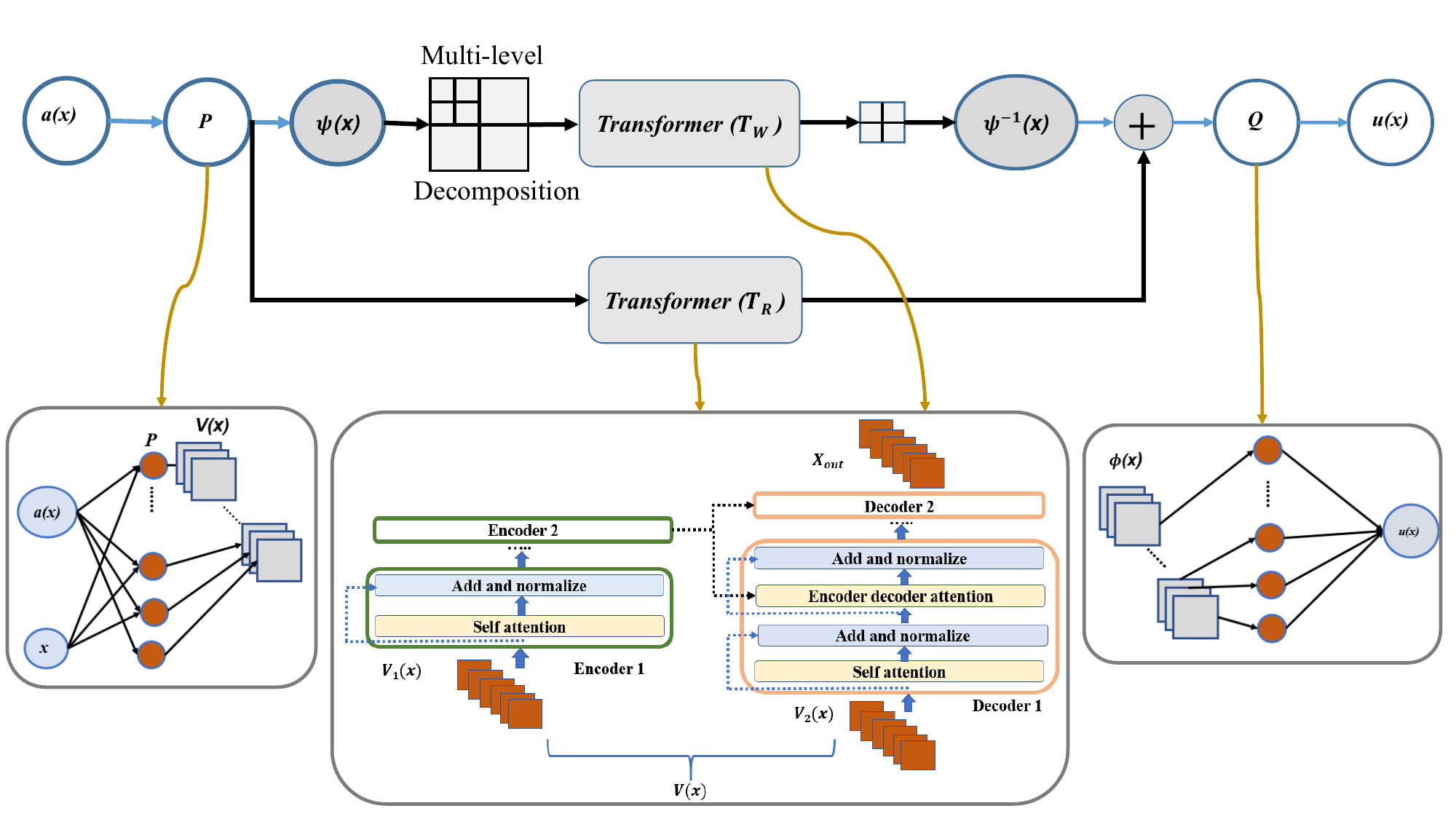}
    \caption{\textbf{A diagrammatic representation of the waveformer architecture}. Firstly, inputs are uplifted by passing through  local transformation $P$. The integral layer consists of two separate branches. In the first branch, a wavelet decomposition is performed on the uplifted images and then is passed though a transformer. In the second branch, the inputs are directly fed to the other transformer. An activation is applied in the   resultant output obtained by summing the outputs of the two branches. Then the outputs are down lifted by passing through the transformation $Q$, which yields the prediction u(x). Here the local transformations $P$ and $Q$ are modeled as fully connected neural networks of one hidden layer.}
    \label{fig:architectur}
\end{figure}

The overall framework of the proposed waveformer is illustrated in \autoref{fig:architectur}. The figure provides the schematic description of the architecture of the proposed framework.
The proposed waveformer is devised as an operator learning paradigm that aims to predict the future spatiotemporal solutions from the initial spatiotemporal response observed over limited time steps. Now, let us consider $\bm{u}_{k:k+n}$ as the solution of the PDE from time-step $k$ to a time-step $k+n$, and initial responses till $k$ time steps $\bm{u}_{0:k}$ with $d_0$ state variables on a given domain discretization of $\Omega$. The discretization on a domain $\Omega$ is done with $D_i$ points in the $i$-th dimension. Specifically, for 1D, $\bm{u}^n \in \mathbb{R}^{d_0 \times D_1}$; for 2D, $\bm{u}^n \in \mathbb{R}^{d_0 \times D_1 \times D_2}$; and for 3D, $\bm{u}^n \in \mathbb{R}^{d_0 \times D_1 \times D_2 \times D_3}$. In the given problem setting, the initial input to the waveformer is given by $\bm a_1(x) = \bm u_{0:k}$  and is trained to predict response, $u_{k+1}$ at the very next time step. Here we note that due to the inherent nature of the transformer which seeks two inputs; one for the encoder and the other for the decoder. Thus we split the initial input, $\bm a^{0}(x)$ into two parts $\bm a^{0_{enc}}(x) = \bm u_{0:k-1}$ and $\bm a^{0_{dec}}_{2}(x) = \bm u_{1:k}$. Following the prediction of the response at ${k+1}^{th}$ time step, we update the splitted inputs such that $\bm a^{1_{enc}}(x)=\{\bm a^{0}_{1:k-1}(x),\bm u_{k}\}$ and $\bm a^{1_{dec}}(x)=\{\bm a^{0}_{2:k}(x),\bm u_{k+1}\}$. Thus, overall input in the next time step can be given as $\bm a^{1}(x)=\{\bm a^{0}_{1:k}(x),\bm u_{k+1}\}$. Similarly, input for the subsequent time step is updated as $\bm a^{2}(x)=\{\bm a^{1}_{1:k}(x),\bm u_{k+2}\}$. Thus, the general updation scheme at $n^{th}$ time step is expressed as:
\begin{equation}
    \bm a^{n}(x)=\{\bm a^{n-1}_{1:k}(x),\bm u_{k+n}\}.
\end{equation}
By recursively applying this training scheme, the output prediction $\bm u_{n}$ at the $n^{th}$ time step can be obtained. An algorithm depicting the forward pass of the waveformer model is given in algorithm \ref{algo:prediction}.

\begin{algorithm}[ht!]
    \caption{Prediction of waveformer through forward passing}\label{algo:prediction}
    \begin{algorithmic}[1]
    \Require{Trained waveformer model: $\mathcal{D}_{\theta \theta}$; Test sample:$a^{0}(x)$; Number of time-steps to predict in future:$\bm m$}
        \State {\textbf{Initial inputs to the model:} $a^{0}(x)=\bm u_{0:k}$ and grid $x$.}
        \For{for i $=1, \ldots$, $\bm m$}
            \State{Stack the input and the grid: $u^{i}_{in} \leftarrow \{a^{i}(x), x\}$.}
            \State{Predict the output:$u_{k+i} \leftarrow \mathcal{D}_{\theta \theta}(u^{i}_{in})$}
            \State{Update the input :$a^{i+1}(x)=\{a^{i}_{1:k}(x),u_{k+i}\}$}
        \EndFor
    \Ensure{Prediction of response from time steps $k$ to $m$, $u_{k:m}$}
    \end{algorithmic}
\end{algorithm}

\section{Numerical examples}\label{sec: Numerical example}
In this section, we present a comprehensive evaluation of the proposed framework using four benchmark problems relevant to various engineering systems. To evaluate the performance, the relative Mean Square Error (MSE) is used. We compare the prediction results of the waveformer with WNO and transformer over the trained region and the extrapolated region in all the cases. We also evaluate the performance of the proposed method with the other popular operator learning paradigms, such as FNO and the Multi-wavelet-based operator in the extrapolated region. A summary of results for different example problems is provided in the tables \autoref{Error_var_samples}, \autoref{Error_prediction1}, \autoref{Error_prediction2}, \autoref{Error_prediction3},\autoref{Error_prediction4} and \autoref{stateofart_prediction}. In regards to the details of the architecture and the training of the proposed method, the waveformer consists of a single wavelet decomposition layer, which is followed by a transformer, where the number of encoder blocks and decoder blocks typically ranges from 1 to 2, tailored to each specific example considered. Specifics of the waveformer architecture and details of baseline models, WNO, and Transformers employed in each numerical example are provided in \autoref{specs_of_arch}. In order to optimize the network parameters, the ADAM optimizer with an initial learning rate varying from $10^{-4}$ to $10^{-3}$, and a weight decay of $10^{-4}$ is used. A learning rate decay of 0.75 is scheduled to enhance the convergence during the training, where the learning rate is reduced by a factor of 0.75 every 20 epochs. While the training of the waveformer is done for 200 epochs, batch sizes vary between 5 and 10 in accordance with the numerical experiment. All the numerical experiments are carried out on a single RTX A5000 GPU with 16 GB of memory.

\begin{table}[ht!]
\caption{\textbf{Architecture specifications of waveformer, WNO, and transformer}. The details of the waveformers, WNOs, and transformers employed for all the presented numerical examples are summarized in the table below.}\label{specs_of_arch}
\centering
\setlength{\tabcolsep}{3pt}
    \begin{tabular}{cc|cccccccc}\hline
        \textbf{Model}&\textbf{Example} & \text { Decomposition} & {Batch}&\text {P} & \text {Q} & \text {Transformer } & \text {Transformer}&{Wavelet} \\
        {(Architecture)} &{}& {levels } & {size} & {} & {} & {encoders} & {decoders}&{}\\ \hline
        \multirow{4}{*}{\textbf{Waveformer}}&\ref{example_1} \text {Example 1}  & 3 &{5}& 80 & 128 & 1 & 2 &{db6}\\
        {}& \ref{example_2} \text {Example 2}& 3 & {2}&80 & 128 & 1 & 2 &{db4}\\
        {}& \ref{example_3} \text {Example 3}& 4 & {5}&40 & 128 & 1 & 2&{db4} \\
        {}& \ref{example_4} \text {Example 4}& 4 &{10}& 40 & 128 & 1 & 2 &{db4}\\
        \hline\hline        
    \end{tabular}
    % \begin{tabular}{cc p{0.7cm}|cccccccc}\hline
    %     \textbf{Framework:}&\textbf{Example}&{}  & \text { Decomposition} & {Batch}&\text {P} & \text {Q} & \text {Transformer } & \text {Transformer}&{Wavelet} \\
    %     {} &{}& &{}{levels } & {size} & {} & {} & {encoders} & {decoders}&{}\\ \hline
    %     \multirow{4}{*}{\textbf{Waveformer}}&\ref{example_1} \text {Example 1}  & &{}3 &{5}& 80 & 128 & 1 & 2 &{db6}\\
    %     {}& \ref{example_2} \text {Example 2}&&{} 3 & {2}&80 & 128 & 1 & 2 &{db4}\\
    %     {}& \ref{example_3} \text {Example 3} &&{} 4 & {5}&40 & 128 & 1 & 2&{db4} \\
    %     {}& \ref{example_4} \text {Example 4} &&{} 4 &{10}& 40 & 128 & 1 & 2 &{db4}\\
    %     \hline\hline        
    % \end{tabular}
    \setlength{\tabcolsep}{4pt}
    \begin{tabular}{cc|cccccccc}
        \textbf{Model}&\textbf{Example}  & \text { Decomposition} & {Batch}&\text {FNN1} & \text {FNN2} & \text {Wavelet} & \text {Wavelet} \\
         {(Architecture)} &{}& {levels } & {size} & {} & {} & {integral layers} & {}\\ \hline
        \multirow{4}{*}{\textbf{WNO}}&\ref{example_1} \text {Example 1}  & 3 &{5}& 80 & 128 & 3 & db6 \\
        {}&\ref{example_2} \text {Example 2}  & 3 & {2}&80 & 128 & 3 & db6 \\
        {}&\ref{example_3} \text {Example 3} & 3 & {5}&40 & 128 & 4 & db4 \\
        {}&\ref{example_4} \text {Example 4}  & 3 &{10}& 40 & 128 & 4 & db4 \\
        \hline\hline
    \end{tabular}
    \setlength{\tabcolsep}{4pt}
    \begin{tabular}{cc|cccccccc}
         \textbf{Model}&\textbf{Example}  &  \text {Transformer} & \text {Transformer} & {Batch}&\text {Number of heads} & \text {Channels}   \\
        {(Architecture)}&{} & {encoders}& {decoders} & {size} & {in attention layer} & {} \\ \hline
        \multirow{4}{*}{\textbf{Transformer}}&\ref{example_1} \text {Example 1}  & 1 &{2}& 5 & 1-3 & 32  \\
        {}&\ref{example_2} \text {Example 2}  & 1 & {2}&2 & 1-3 & 32  \\
        {}&\ref{example_3} \text {Example 3} & 1 & {2} & 5 & 1-3 & 32  \\
        {}&\ref{example_4} \text {Example 4} & 1 &{2}& 10 & 1-3& 40 \\
        \hline
    \end{tabular}
\end{table}

\subsection{Example 1: Burgers’ diffusion dynamics}\label{example_1}
As the first example, we investigate the 1D Burgers' equation, which is widely employed to describe various phenomena such as wave formation, turbulence, fluid flows in fluid mechanics, gas dynamics, and traffic flow \cite{kutluay1999numerical, wazwaz2002partial}. The one-dimensional Burgers' equation with periodic boundary conditions is mathematically defined as follows:
\begin{equation}
    \begin{aligned}
    \frac{\partial u}{\partial t}+u \frac{\partial u}{\partial x} & =\nu \frac{\partial^2 u}{\partial x^2}, & & x \in(0,1), t \in(0,1] \\
    u(x=0, t) & = u(x=1, t) =0, & & x \in(0,1), t \in(0,1] \\
    u(x, 0) & = u_0(x), & & x \in(0,1).
    \end{aligned}
\end{equation}
where $\nu$ represents the viscosity of the flow and is fixed to a value of $0.1$ in this case. \textbf{Data generation:} Initial conditions are chosen of the form $u_0(x) = \cos(\zeta \pi x) + \sin(\eta \pi x)$, while multiple samples of initial conditions are generated by varying the parameters $\zeta$ and $\eta$ such that  $\zeta \sim \operatorname{Unif}(0.5, 1.5)$ and $\eta \sim \operatorname{Unif}(0.5, 1.5)$. The objective here is to learn the operator $\mathcal{D}$ which maps the spatiotemporal response of initial $k$ times steps, $u_{0:k}(x)$ to the spatiotemporal solution at the next $n$ time steps $u_{k:k+n}(x)$. $\mathcal{D}: u_{0:k}(x) \mapsto u_{k:k+n}(x)$. The ground truth datasets are generated using a MATLAB PDE solver, where the spatial resolution is chosen to be $60$. The $k$ and $n$ values used in the problems are $51$ and $60$, respectively. The model is trained with the randomly generated instances of number $320$, and the accuracy of the model is evaluated on $20$ unseen realizations. A validation set of realizations is employed to determine the optimal hyperparameters of the model.

\begin{figure}[!ht]
    \centering{
    \includegraphics[width=0.5\textwidth]{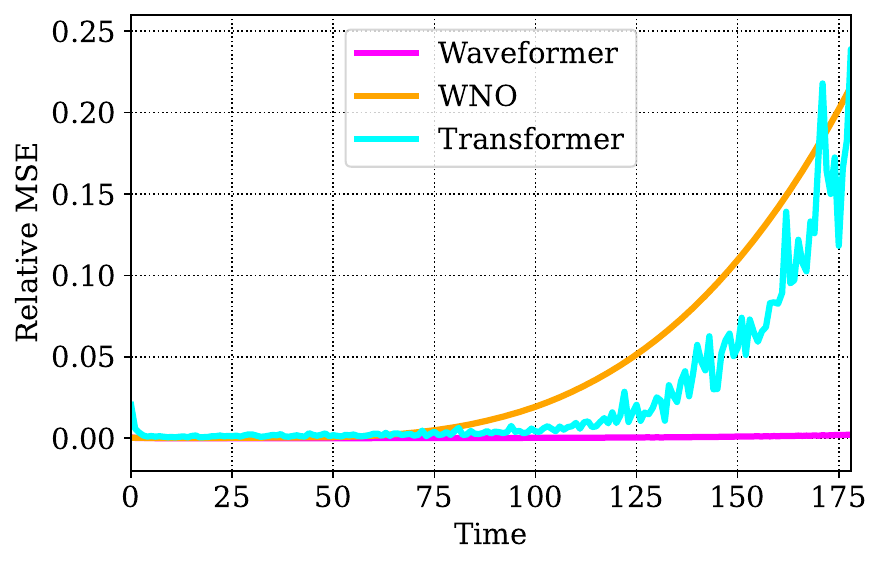}}
    \caption{Variation of prediction error (Relative MSE) of the Waveformer, WNO, and the transformer with time steps in the Burgers’ diffusion dynamics example}\label{RMSE_t_burger}
\end{figure}

\begin{figure}[!ht]
    \centering{
    \includegraphics[width=0.8\textwidth]{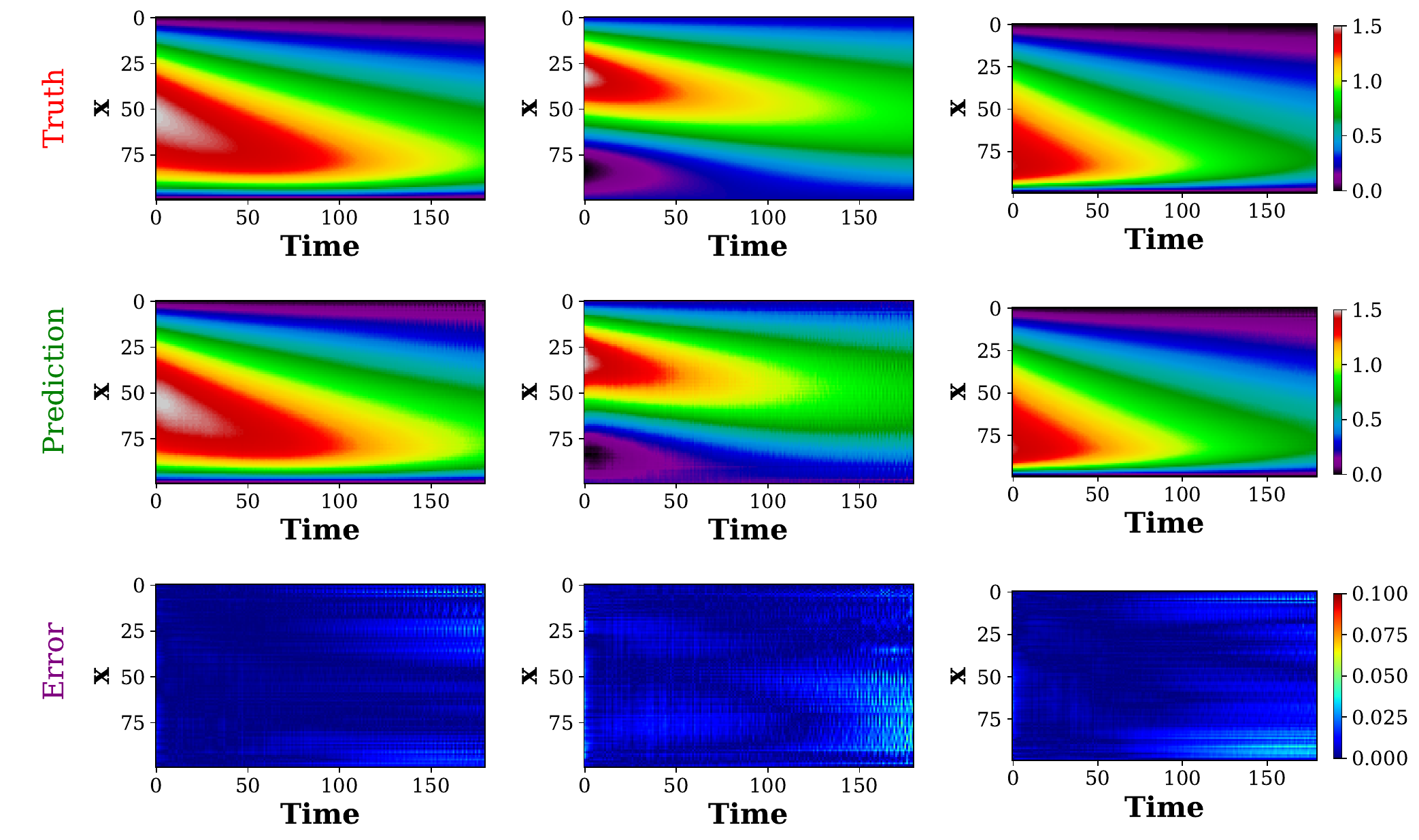}}
    \caption{Predictions of three different test samples of Burger's equation employing waveformer. (Top to bottom) ground truth solution, waveformer prediction, and the error plot with $L_1$ error.}\label{fig:burger}
\end{figure}

\textbf{Results}: The prediction results provided in the \autoref{Error_prediction1} and \autoref{stateofart_prediction} indicate the efficacy of the waveformer in predicting the time marched response of the system. The \autoref{RMSE_t_burger} shows the variation of the prediction error of the proposed approach, WNo, and transformer with time. While the waveformer yields excellent results even in the extrapolated time regime (relative MSE of $6\times 10^{-4}$), WNO and Transformer show a significant deviation from the ground truth solution. Contour plots of the predicted spatiotemporal response in comparison with the ground truth solution are shown in \autoref{fig:burger}

\begin{table}[ht!]
    \centering
    \caption{\textbf{Results of the prediction error of the waveformer for the Burger diffusion dynamics}. The prediction error of the optimized model after training is computed over the test data.\\}
    \label{Error_prediction1}
    \begin{tabular}{lccccc} 
        \toprule
        {\textbf{Time range}}&\multicolumn{3}{c}{\textbf{Relative MSE}} \\ \cmidrule{2-4}
        \textbf{(in time steps)}& \textbf{Waveformer} & \textbf{WNO} & \textbf{Transformer}  \\ \midrule
        \textbf{Trained region [0 - 60)} & $ 5.76\times 10^{-5} $ & $4\times 10^{-4}$ & $1.9\times 10^{-3}$ \\
       
        \textbf{Extrapolated region [60 - 180)}& $6\times 10^{-4} $ & $ 6.5\times 10^{-2}$ & $3.9\times 10^{-2}$\\
        \bottomrule  
    \end{tabular}
\end{table}

\subsection{Example 2: Kurumoto Sivasinsky Equation}\label{example_2}
\begin{figure}[ht!]
\centering
    \centering
    \subfigure[sample-1]{\includegraphics[width=0.9\textwidth]{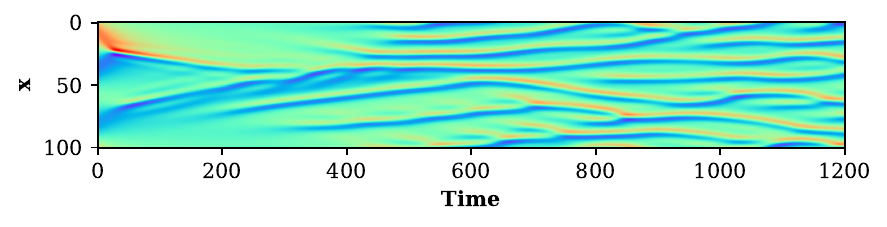}} 
    \subfigure[sample-2]{ 
    \includegraphics[width= 0.9\textwidth]{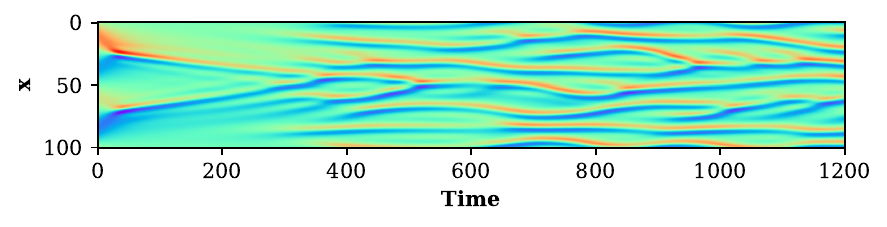}}
    \caption{The simulated ground-truth data using a spectral ETDRK4 scheme and implemented in MATLAB. Visualization of Two samples of the generated ground truth solutions are shown above.}
    \label{fig:samplesol}
\end{figure}

\begin{figure}[!ht]
    \centering{
    \includegraphics[width=0.8\textwidth]{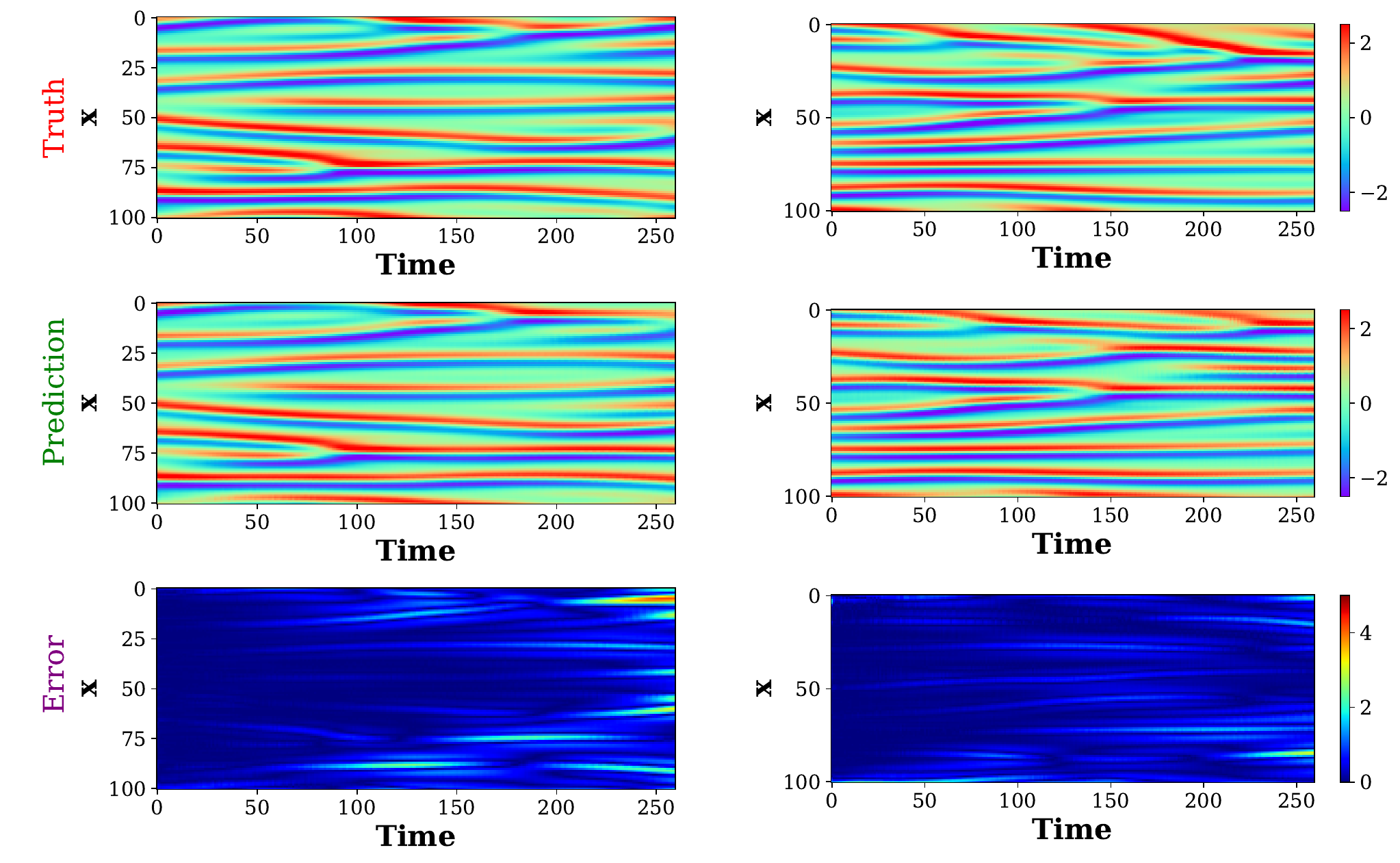}}
    \caption{Predictions of two different test samples of K-S equation employing waveformer. (Top to bottom) ground truth solution, waveformer prediction, and the error plot with $L_1$ error.}\label{fig:K-S equation}
\end{figure}

We now consider the Kurumoto Sivasinsky equation (K-S equation) as our second numerical example. The K-S equation is a fourth-order 1-D time-dependent PDE commonly used to model chemical phase turbulence, plasma ion instabilities and flame front instabilities. The problem is relatively challenging as the PDE describes the chaotic dynamics of weakly turbulent systems. The K-S equation with periodic homogeneous Neumann boundary conditions can be described as follows:
\begin{equation}
    \begin{aligned}
    \frac{\partial u}{\partial t}+u \frac{\partial u}{\partial x}+\frac{\partial^2 u}{\partial x^2}+\nu \frac{\partial^4 u}{\partial x^4} & =0, & & x \in(0,L), t \in(0,1] \\
    u(x=0, t) & =u(x=1, t), & & x \in(0,L), t \in(0,1] \\
    u(x, 0) & =u_0(x), & & x \in(0,L).
    \end{aligned}
\end{equation}
where the parameter $\nu$ represents the viscosity parameter and is set to a value $\nu=1$. \textbf{Data generation:} The size of the domain is chosen to be $L=[0,22\pi]$, while it is discretized into $101$ points over the space. For obtaining the temporal response, we use the time-step of $\delta t = 0.1$. The random initial conditions are generated such that \cite{geneva2020modeling}:
\begin{equation}
\begin{aligned}
& u(x, 0)=2 c \frac{\beta(x)-\min _x \beta(x)}{\max _x \beta(x)-\min _x \beta(x)}-c, \quad w(x)=\sum_{n=1}^3 \frac{\lambda_n}{n} \sin \left(\frac{n \pi x}{l}+ b\right), \\
& \lambda_n=[1, \mathcal{N}(0,2), 1], \quad b =2 \pi \operatorname{Unif}[0,1], \quad c=\mathcal{N}(0,0.5)+c_0, \\
& l=L /\left(2 k_0\right), \quad k_0=\lfloor L /(2 \pi \sqrt{2})+0.5\rfloor,
\end{aligned}
\end{equation}
The constant $c_0$ is set to a value of $2.5$, which is estimated as \textit{a priori} mean of the amplitude. The parameter $k_0$ represents the number of unstable modes and is estimated based on the domain length, using the formula $k_0 =\lfloor L /(2 \pi \sqrt{2})+0.5\rfloor$ \cite{cvitanovic2010state}. For simulating the ground truth solution, a spectral ETDRK4 scheme is employed. Moreover, training data is obtained by omitting the initial transient state response of $700$ time steps. While the training set contains 180 samples, and the test set consists of 20 samples. Two such generated samples are shown in \autoref{fig:samplesol}.
Similar to the previous example, we seek an operator  $\mathcal{D}: u_{0:k}(x) \mapsto u_{k:k+n}(x)$. The values of $k$ and $n$ are set to be $k= 51$ and $n = 60$. Further, the model is trained for 180 randomly generated instances and is validated on 20 unseen samples. A visual depiction of  
the predicted spatiotemporal response results are provided in \autoref{fig:K-S equation}.

\begin{figure}[!ht]
    \centering{
    \includegraphics[width=0.5\textwidth]{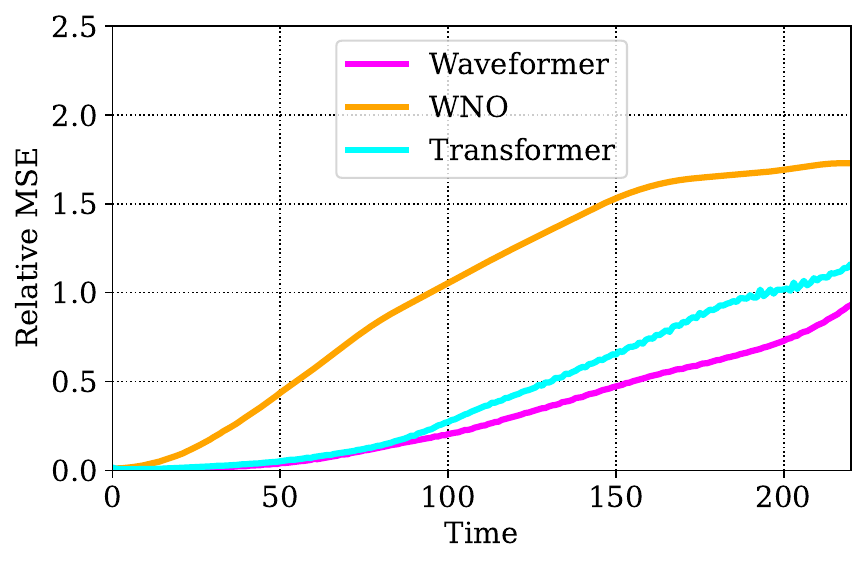}}
    \caption{Variation of prediction error (Relative MSE) of the Waveformer, WNO, and the transformer with time steps in the K-S equation example.}\label{RMSE_t_K-S_equation}
\end{figure}

\textbf{Results}: The prediction results presented in \autoref{Error_prediction2} and \autoref{RMSE_t_K-S_equation} reinforce the effectiveness of the waveformer in predicting accurate dynamics of the system. While the Waveformer outperforms the WNO and Transformer in predicting the future response, we note that the trained transformer yields a comparable result as that of the waveformer in the training region. It is also observed from the \autoref{stateofart_prediction} that Waveformer incurs the least prediction error among the state-of-the-art methods.
\begin{table}[ht!]
    \centering
    \caption{\textbf{Results of the prediction error of the waveformer for the Kurumoto Sivasinsky Equation}. The prediction error of the optimized model, after training, is computed over the test data.\\}
    \label{Error_prediction2}
    \begin{tabular}{lccccc} 
        \toprule
        {\textbf{Time range}}&\multicolumn{3}{c}{\textbf{Relative MSE}} \\ \cmidrule{2-4}
        \textbf{(in time steps)}& \textbf{Waveformer} & \textbf{WNO} & \textbf{Transformer}  \\ \midrule
        \textbf{Trained region [0 - 60)} & $ 0.0203 $ & $0.2157 $ & $0.0279$ \\

        \textbf{Extrapolated region [60 - 260)}& $0.5635 $ & $ 1.4067$ & $0.7185$\\
        \bottomrule  
    \end{tabular}
\end{table}

\subsection{Example 3:2-D Allen-Cahn equation}\label{example_3}
The Allen-Cahn equation is a well-known partial differential equation often used for modeling reaction-diffusion phenomena in various scientific domains \cite{lord2014introduction,ma2017numerical}. The mathematical form of the Allen-Cahn equation is described as:
\begin{equation}
\begin{aligned}
\frac{\partial u(x, y, t)}{\partial t} &= \epsilon \Delta u(x, y, t) + u(x, y, t) - u(x, y, t)^3,\\
u(x, y, 0) & = u_0(x, y), && (x, y) \in (0, 1)
\end{aligned}
\end{equation}
Here, the parameter $\epsilon$ represents the viscosity coefficient, and for the present case, we set $\epsilon = 1 \times 10^{-3}$. The equation is defined on a periodic spatial domain. \textbf{Data generation:} To generate the initial conditions, the GRF with the following kernel is utilized:
\begin{equation}
    \mathcal{K}(x, y)=\tau^{(\alpha-1)}\left(\pi^2\left(x^2+y^2\right)+\tau^2\right)^{\frac{\pi}{2}}.
\end{equation}
Here the parameters for the kernel are chosen as $\tau=15$ and $\alpha=1$. Training and testing datasets are simulated using the spectral Galerkin method \cite{lord2014introduction}. Here we train the proposed framework to learn the operator  $\mathcal{D}: u_{0:k}(x) \mapsto u_{k:k+n}(x)$ such that the value of $k$ and $n$ are chosen for the training are $10$ and $25$ respectively. While the training set contains 300 samples, the testing data set contains 20 samples.

\begin{figure}[!ht]
    \centering{
    \includegraphics[width=0.5\textwidth]{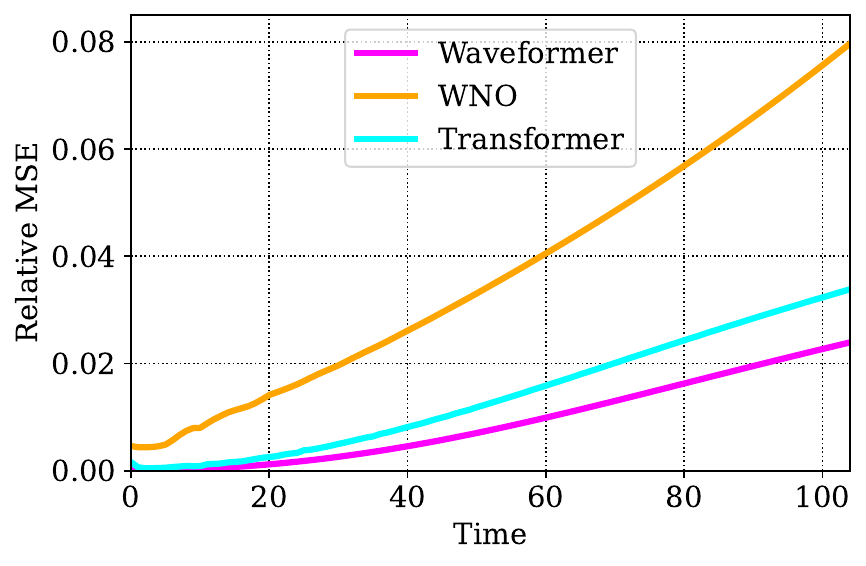}}
    \caption{Variation of prediction error (Relative MSE) of the Waveformer, WNO, and the transformer with time steps in the Allen-Cahn example.}\label{RMSE_t_allencahn}
\end{figure}

\begin{figure}[!ht]
    \centering{
    \includegraphics[width= 1\textwidth]{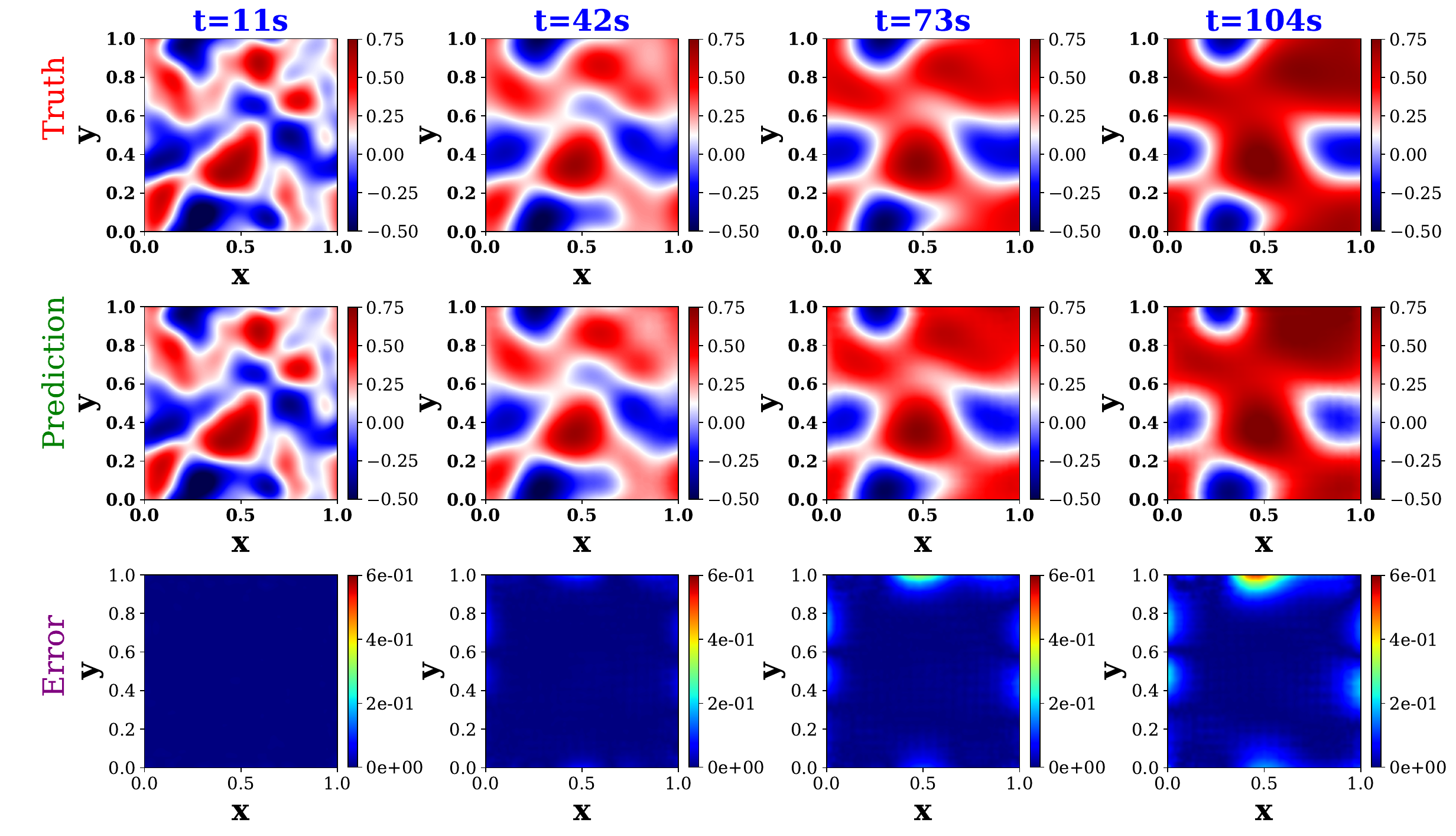}}
    \caption{Predictions of a single test sample of 2-D Allen-Cahn equation at 
    4 different time instances employing waveformer. (Top to bottom) ground truth solution, waveformer prediction, and the error plot with $L_1$ error.}\label{fig:Allencahn}
\end{figure}

\textbf{Results}: The prediction results depicted in \autoref{Error_prediction3},\autoref{stateofart_prediction} and \autoref{RMSE_t_allencahn} demonstrate the superior performance of the waveformer model compared to WNO and Transformer. It is worth noting that this problem exemplifies the scalability of the waveformer framework, as it successfully handles higher-dimensional problems beyond the scope of previous 1-D dynamical systems. A visualization of results is provided in \autoref{fig:Allencahn}, where the predicted results of a test sample at different time instances are shown in comparison with the target solution.
\begin{table}[ht!]
    \centering
    \caption{\textbf{Results of the prediction error of the waveformer for the Allen-Cahn equation}. The prediction error of the optimized model, after training, is computed over the test data.\\}
    \label{Error_prediction3}
    \begin{tabular}{lcccc} 
        \toprule
        {\textbf{Time range}}&\multicolumn{3}{c}{\textbf{Relative MSE}} \\ \cmidrule{2-5}
        \textbf{(in time steps)}& \textbf{Waveformer} & \textbf{WNO} & \textbf{Transformer}  \\ \midrule
        \textbf{Trained region [0 - 25)} & $ 0.0012 $ & $0.0124$ & $0.0025$ \\
        \textbf{Extrapolated region [25- 105)}& $0.0131$ & $ 0.0491$ & $0.0199$\\
        \bottomrule  
    \end{tabular}
\end{table}

\subsection{2D time-dependent Navier–Stokes equation}\label{example_4}
The Navier-Stokes equation is a well-known second-order nonlinear parabolic partial differential equation (PDE). The equation describes the conservation of momentum, and it has a wide range of applications, including fluid dynamics, analysis of airflow around airplane wings, modeling ocean currents, and investigating thermodynamic processes. The governing equations for the 2D incompressible Navier-Stokes equation in the vorticity-velocity form are provided as follows:
\begin{equation}
\begin{aligned}
\frac{\partial \omega(x, y, t)}{\partial t}+u(x, y, t) \cdot \nabla \omega(x, y, t) & =v \Delta \omega(x, y, t)+f(x, y), & & x, y \in(0,1), t \in(0, T] \\
\nabla \cdot u(x, y, t) & =0, & & x, y \in(0,1), t \in[0, T] \\
\omega(x, y, 0) & =\omega_0(x, y), & & x, y \in(0,1)
\end{aligned}
\end{equation}
Here, the scalar parameter $v$ represents the fluid viscosity, and $f(x, y)$ is the source function. The variables $u(x, y, t)$ and $\omega(x, y, t)$ denote the velocity and vorticity fields of the fluid, respectively. The initial vorticity field $\omega(x, y, 0)$ is given by $\omega_0(x, y)$ for $x, y \in (0, 1)$. The viscosity value is set as $v = 10^{-3}$. This study aims to learn an operator that maps the vorticity fields of initial $k$ time steps to vorticity fields from $k+n$ through forward passing. The mapping is defined such that $\mathcal{D}: u_{0:k}(x) \mapsto u_{k:k+n}(x)$ and value of $k$ and $n$ chosen for the training are $14$ and $21$ respectively. While the training set contains 420 samples, the testing data set contains 20 samples.
\textbf{Data generation}
A GRF  is used to generate the random initial vorticity field $\omega_0(x, y)$ such that $\omega(x, y, 0) = \mathcal{N}\left(0,7^{3 / 2}(-\Delta+49 I)^{-2.5}\right)$. The source function $f(x, y)$ in this case remains constant and is given by ${0.1}(\sin (2 \pi(x+y)) + \cos (2 \pi(x+y)))$. The Crank-Nicolson scheme with a time-step of $10^{-4}$ is employed for the time integration. Here, we note that to train the neural network, the solution data is stacked at intervals of $t = 0.5$ second.
Further, the initial $50$ time-step responses are omitted to enhance the training. The spatial resolution of the vorticity fields is fixed to be $64 \times 64$ \cite{li2020fourier}. The same spatial resolution is used for training and testing. 

\begin{figure}[!ht]
    \centering{
    \includegraphics[width=0.5\textwidth]{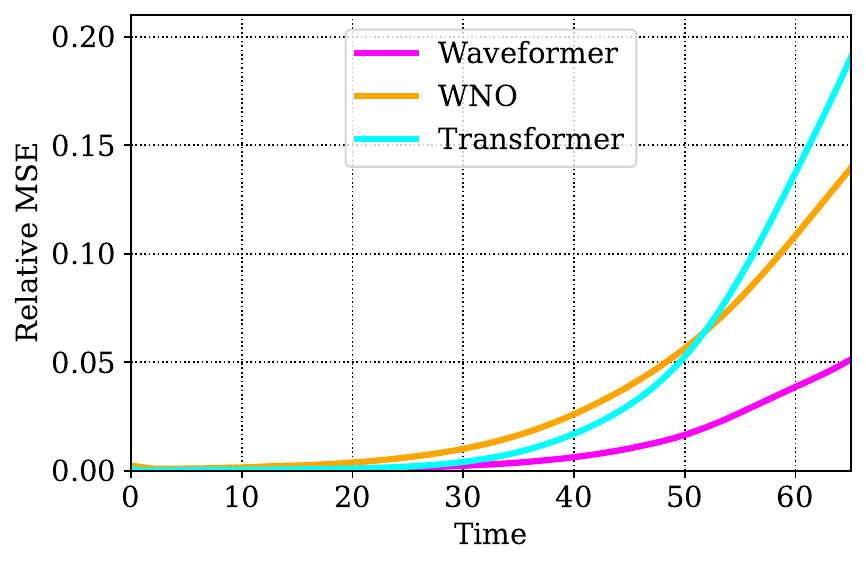}}
    \caption{Variation of prediction error (Relative MSE) of the Waveformer, WNO, and the transformer with time steps in the Navier-stokes equation example.}\label{RMSE_t_navier_stokes}
\end{figure}

\begin{figure}[!ht]
    \centering{
    \includegraphics[width= 1\textwidth]{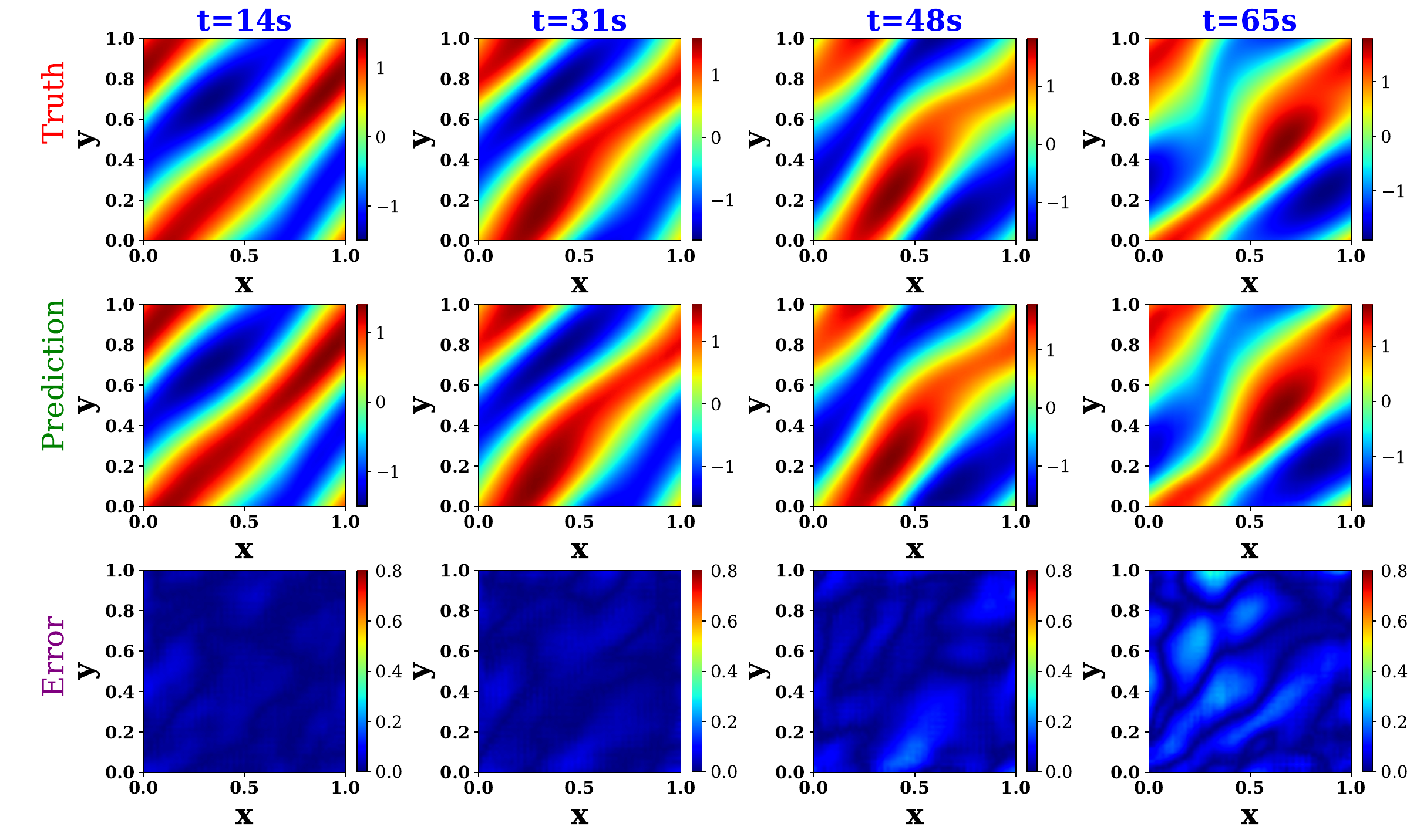}}
    \caption{Predictions obtained by waveformer for a single test sample of 2-D time-dependent Navier-Stokes equation at 4 different time instances. (Top to bottom) ground truth solution, waveformer prediction, and  prediction error.}\label{fig:Navierstokes}
\end{figure}

\textbf{Results:} The prediction results for the vorticity field at $t = \{14,66\}$ along with the mean prediction errors are summarized in \autoref{Error_prediction4}. While, from the \autoref{Error_prediction4}, it can be observed from the results that the waveformer model outperforms the WNO and Transformer both in the training interval and the extrapolated regime, the \autoref{stateofart_prediction} showcases the performance of in the extrapolated region in comparison with the state of the art operators. It is evident from the result depicted in the \autoref{RMSE_t_navier_stokes} that with time, the prediction error increases in all the presented methods; however, the prediction of waveformer has significantly less deviation from the ground truth compared to that of the WNo, and the transformer. Furthermore, a visualization of the results is illustrated in \autoref{fig:Navierstokes}
\begin{table}[ht!]
    \centering
    \caption{\textbf{Results of the prediction error of the waveformer for the Navier-Stokes equation}. The prediction error of the optimized model is computed over the test data.\\}
    \label{Error_prediction4}
    \begin{tabular}{lccccc} 
        \toprule
        {\textbf{Time range}}&\multicolumn{3}{c}{\textbf{Relative MSE}} \\ \cmidrule{2-4}
        \textbf{(in time steps)}& \textbf{Waveformer} & \textbf{WNO} & \textbf{Transformer}  \\ \midrule
        \textbf{Trained region [0 - 21)} & $4\times 10^{-4} $ & $1.9 \times 10^{-3}$ & $5\times 10^{-4}$ \\
        \textbf{Extrapolated region [21- 66)}& $0.0143$ & $ 0.0451$ & $0.0471$\\
        \bottomrule  
    \end{tabular}
\end{table}

\begin{table}[ht!]
    \centering
    \caption{\textbf{Results of the prediction error of waveformer with varying training samples for the first three illustrated examples}. The prediction error corresponds to the relative Mean Square Error (MSE) of the optimized model after training, computed over the test data. $N_s$ represents the number of training samples.\\}
    \label{Error_var_samples}
    \begin{tabular}{lcccccc} 
        \toprule
        {\textbf{Numerical example}} &{}& \multicolumn{3}{c}{\textbf{Prediction error of waveformer}}&{}
        \\ \midrule
         {Burger's equation} &\textbf{Relative MSE} & $4\times 10^{-3} $ & $1.7\times10^{-3} $ & $1.3\times 10^{-3}$&{$5.87\times 10^{-5}$}\\
         {} & $\mathbf{N_s}$ & {60} & {160} & {240} & {320}\\
        % \hdashline
         {K-S equation } & \textbf{Relative MSE} & $0.1023$ & $0.0557$ & $0.0342$&{$0.0203$}\\
        {} & $\mathbf{N_s}$ & {50} & {100} & {140} & {180}\\
        % \hdashline
         {Allen-Cahn equation} & \textbf{Relative MSE} & $2.6 \times 10^{-3}$ & $ 1.7 \times 10^{-3}$& $ 5\times 10^{-4}$ & $ 5 \times 10^{-4}$\\
         {} & $\mathbf{N_s}$ & {60} &{100} & {200} & {300}\\
        \bottomrule  
    \end{tabular}
\end{table}

\begin{table}[ht!]
\centering
\caption{\textbf{Results of the prediction error of the waveformer in compassion with the state-of-the-art operator learning methods}. The prediction error (relative MSE) of the waveformer, along with the operator learning methods, including Multiwavelet-based Operator(MWT), FNO, and WNO in the extrapolated region are summarized below}
\label{stateofart_prediction}
    \begin{tabular}{l|c|ccccccc}
         \toprule 
         \text { Numerical example }  &  \text {Extrapolated} & \text {Waveformer} & {WNO}&\text {MWT \cite{gupta2021multiwavelet}} & \text {FNO \cite{li2020fourier}} &{FNO-3d \cite{li2020fourier}}  \\
        \text {}  & {time range}& {} & {} & {-} & {} &{-}\\  \toprule
        \text {Burgers’ diffusion }  & [60-180] & {$6\times 10^{-4}$} & {$6.5\times 10^{-2}$} & {-} & {$1\times 10^{-3}$} &{-}  \\
        {dynamics}&&&&&\\
        \text {Kurumoto-Sivasinsky}  &[60-200] & 0.3672 & 1.2815 & 1.4936 & 0.4111 &{}   \\
        {equation}&&&&&\\
        \text {Allen-Cahn equation} & [25-106] & 0.0131 & 0.0491 & 0.0286 & - & - \\
        \text {Navier-Stokes equation} & [21-66]] & 0.0143 & 0.0451 & 0.1109 & 0.02 & 0.018 \\
        \hline
    \end{tabular}
\end{table}

\section{Conclusions}\label{sec:conclusion}
In the present work, we propose waveformer, a novel operator learning approach for learning solutions of time-dependent parametric partial differential equations. The proposed waveformer exploits wavelet transform to capture the spatial multi-scale behavior of the solution field and transformers for capturing the long-term temporal dependencies. While transformers and self-attention models have shown great promise in Natural Language Processing (NLP) tasks, their application in other domains is still evolving, especially for modeling physical systems. The waveformer leverages wavelet transformations to map inputs from the physical space to the spectral domain, which enables the spatial and frequency localization of properties inherent in wavelets.
Thus wavelet transformation act as a suitable embedding for the physical states. The architecture of the proposed waveformer is inspired by the neural operator theory. The amalgamation of transformers and wavelets in the waveformer framework offers a promising operator learning framework that can be applied to complex problems transcending the realm of NLP. The results of the numerical experiment affirm the efficacy of waveformer in modeling physical systems. It is worth noting that the waveformer framework exhibits scalability, allowing it to handle increasingly complex systems. The ability of the waveformer to capture complex dynamics, handle multi-scale behaviors, and learn from data makes it a versatile tool for modeling and prediction in various scientific, engineering, and societal domains. A few examples of the diverse range of applications where the waveformer framework can be employed include weather and climate modeling,  predicting fluid flow phenomena, such as airflow around objects, water currents, and turbulence, and modeling material behavior, such as phase transitions, crystal growth, and diffusion processes. It is important to acknowledge that while the waveformer is advantageous in modeling dynamic partial differential equations (PDEs), the reliance on large amounts of training data can pose challenges, especially considering the expensive nature of data generation. Therefore, future research endeavors should focus on incorporating the underlying physics of the system into the model, thereby reducing the data requirements and enhancing the efficiency and applicability of the approach.

\section*{Acknowledgements}
NN acknowledges the financial support received from the Ministry of Education (MoE), India, in the form of the Prime Minister's Research Fellowship (PMRF). SC acknowledges the financial support received from Science and Engineering Research Board via grant no. SRG/2021/000467, Ministry of Port and Shipping via letter no. ST-14011/74/MT (356529) and seed grant received from IIT Delhi

% \bibliographystyle{ieeetr}
% \bibliography{references}  %%% Uncomment this line and comment out the ``thebibliography'' section below to use the external .bib file (using bibtex).

\end{document}